%% file: neurips_2023.tex
\documentclass{article}

\usepackage[preprint]{neurips_2023}
\usepackage{neurips_2023}
\usepackage[utf8]{inputenc} % allow utf-8 input
\usepackage[T1]{fontenc}    % use 8-bit T1 fonts
\usepackage{hyperref}       % hyperlinks
\usepackage{url}            % simple URL typesetting
\usepackage{booktabs}       % professional-quality tables
\usepackage{amsfonts}       % blackboard math symbols
\usepackage{nicefrac}       % compact symbols for 1/2, etc.
\usepackage{microtype}      % microtypography
\usepackage{xcolor}         % colors
\usepackage{graphicx}
\usepackage{subcaption}

\title{Revealing Model Biases: Assessing Deep Neural Networks via Recovered Sample Analysis}

\author{%
  Mohammad Mahdi Mehmanchi \thanks{Equal contribution}\\
  University of Tehran\\
  \texttt{mahdi.mehmanchi@ut.ac.ir} \\
  \And
   Mahbod Nouri \thanks{Equal contribution}\\
   University of Bremen \\
  \texttt{mahbod@uni-bremen.de} \\
   \AND
   Mohammad Sabokrou$^{1,2}$ \thanks{Corresponding author} \\
   $^1$Okinawa Institute of Science and Technology (OIST)  \\
$^2$Institute For Research In Fundamental Sciences (IPM) \\
  \texttt{mohammad.sabokrou@oist.jP} \\
}

\begin{document}
\maketitle
\begin{abstract}
\input{Abstract}
\end{abstract}
% \begin{bibunit}[plainnat]

\input{Intro}
\input{Related_work}
\input{pre}
\input{new_proposed}

\input{Results}
\input{Discussion}

\input{Conclusion}
\bibliography{neurips_2023}
\bibliographystyle{plainnat}
% \putbib[references]
% \end{bibunit}

% All headings should be lower case (except for first word and proper nouns),
% flush left, and bold.

% First-level headings should be in 12-point type.

% \subsection{Headings: Preliminaries}

% \begin{ack}
% Use unnumbered first level headings for the acknowledgments. All acknowledgments
% More information about this disclosure can be found at: \url{https://neurips.cc/Conferences/2023/PaperInformation/FundingDisclosure}.

% Do {\bf not} include this section in the anonymized submission, only in the final paper. You can use the \texttt{ack} environment provided in the style file to autmoatically hide this section in the anonymized submission.
% \end{ack}

%%%%%%%%%%%%%%%%%%%%%%%%%%%%%%%%%%%%%%%%%%%%%%%%%%%%%%%%%%%%
% \bibliography{references}

\appendix
% \section{Appendix}
% \begin{bibunit}[plainnat]
\section{Hyperparameter details}
There are four hyperparameters for the reconstruction algorithm. For all experiments conducted on CIFAR10 \citep{krizhevsky2009learning} and Tiny ImageNet \citep{le2015tiny} datasets, we utilized Weights \& Biases \citep{biewald2020experiment} to find the best set of hyperparameters. Our criterion was the mean SSIM of best 10 reconstructed images. As suggested in \citep{haim2022reconstructing}, we conducted a random grid search by sampling parameters from the mentioned distributions in that paper. 
However, in both CIFAR10 and Tiny ImageNet experiments, the optimized hyperparameters suggested in \cite{haim2022reconstructing} was also optimal for our experiments. 
We also used the suggested hyperparameters for our MNIST \citep{6296535} experiments.

\section{Additional Experiments on Tiny ImageNet}
Alongside the experiments conducted on CIFAR10 and MNIST, we also experimented on the Tiny ImageNet dataset. In this experiment, we defined food and animals as our two classes of interest. Specifically, we combined five related food categories and five related animal categories. As in our previous experiments, we used a training set of 100 images, comprising 50 images from each class. Moreover, we introduced two different levels of bias difficulty in our training data to construct models \(\mathcal{M}_1\) and \(\mathcal{M}_3\) (for a detailed explanation of these models see the proposed method section of the paper). 
In table \ref{tab:table1}, mean SSIM of best 10 reconstructed samples as well as the test accuracy on clean test set are reported. As we can see, the mean SSIM of the best 10 reconstructed samples is lower for model \(\mathcal{M}_1\) than for other models. We also notice that model \(\mathcal{M}_1\) has the worst performance (in terms of test accuracy) compared to other models. Therefore, these experiments also validate our main idea that by reconstructing the training data, we can assess to what extent the model has learned the main features of the training data.
On a side note, we see that the mean SSIM of model \(\mathcal{M}_3\) is slightly higher than the model \(\mathcal{M}_{base}\) , while the test accuracy of model \(\mathcal{M}_{base}\) is slightly better than model \(\mathcal{M}_3\).
This is probably because of the fact that reconstructing the trigger squares are easy, leading to an increase in the SSIM value for model \(\mathcal{M}_3\) compared to the model without trigger bias (\(\mathcal{M}_{base}\)).

\begin{figure}
\centering
\begin{subfigure}{0.4\textwidth}
    \includegraphics[width=\textwidth]{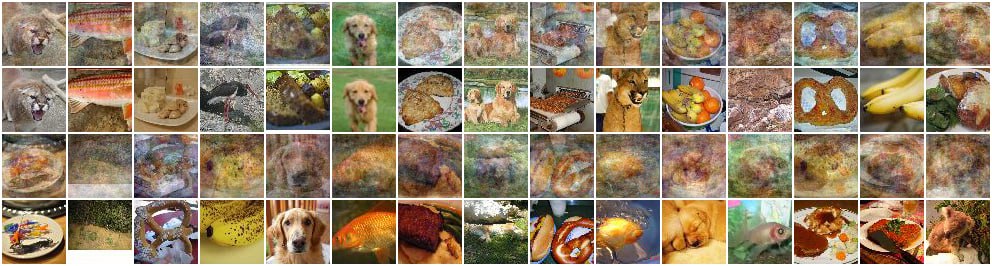}
    \caption{Model \(\mathcal{M}_{base}\)}
    \label{fig:first}
\end{subfigure}
\hfill
\begin{subfigure}{0.4\textwidth}
    \includegraphics[width=\textwidth]{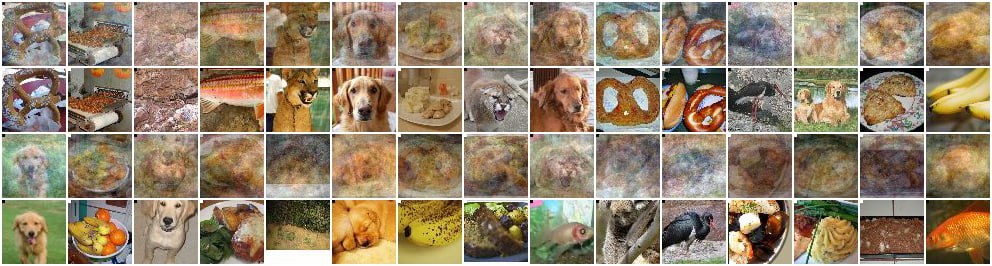}
    \caption{Model \(\mathcal{M}_{1}\)}
    \label{fig:second}
\end{subfigure}
\hfill
\begin{subfigure}{0.4\textwidth}
    \includegraphics[width=\textwidth]{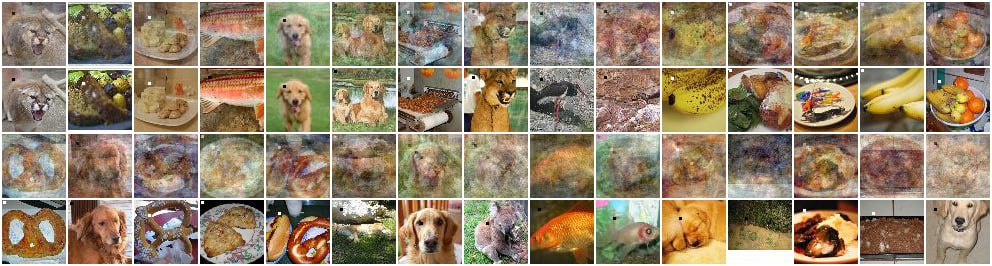}
    \caption{Model \(\mathcal{M}_{3}\)}
    \label{fig:third}
\end{subfigure}
        
\caption{Top 30 reconstructed samples for different models trained on Tiny Imagenet dataset with different bias levels. Odd rows show reconstructed samples and even rows show the corresponding training samples.}
\label{fig:fig1}
\end{figure}

The figure depicted in Figure \ref{fig:fig1} displays the top 30 reconstructed images for each model. It is apparent that the reconstructed samples of model \(\mathcal{M}_1\) are inferior to those generated by the other two models. This finding also indicates that model \(\mathcal{M}_1\) has lower quality compared to the other two models.

\begin{table}
  \caption{Test accuracy of different models trained on 100 Tiny Imagenet images for a binary classification task as well as the average SSIM of best 10 reconstructed images from the weights of the trained model.}
  \label{tab:table1}
  \centering
  \begin{tabular}{lll}
    \toprule
    % \multicolumn{2}{c}{Part}                   \\
    % \cmidrule(r){1-2}
    Model  & Test accuracy    & Mean best 10 SSIMs \\
    \midrule
    \(\mathcal{M}_{base}\) & 0.692   & 0.68    \\
    \(\mathcal{M}_{1}\) & 0.65 & 0.635   \\
    \(\mathcal{M}_{3}\) & 0.682  & 0.698     \\
    \bottomrule
  \end{tabular}
\end{table}

% Bibliography
% \bibliography{sample}

% \bibliographystyle{plainnat}
% \putbib[sample]
% \end{bibunit}

\end{document}

%% file: Abstract.tex
This paper proposes a straightforward and cost-effective approach to assess whether a deep neural network (DNN) relies on the primary concepts of training samples or simply learns discriminative, yet simple and irrelevant features that can differentiate between classes. The paper highlights that DNNs, as discriminative classifiers, often find the simplest features to discriminate between classes, leading to a potential bias towards irrelevant features and sometimes missing generalization. While a generalization test is one way to evaluate a trained model's performance, it can be costly and may not cover all scenarios to ensure that the model has learned the primary concepts. Furthermore, even after conducting a generalization test, identifying bias in the model may not be possible. Here, the paper proposes a method that involves recovering samples from the parameters of the trained model and analyzing the reconstruction quality. We believe that if the model's weights are optimized to discriminate based on some features, these features will be reflected in the reconstructed samples. If the recovered samples contain the primary concepts of the training data, it can be concluded that the model has learned the essential and determining features. On the other hand, if the recovered samples contain irrelevant features, it can be concluded that the model is biased towards these features. The proposed method does not require any test or generalization samples, only the parameters of the trained model and the training data that lie on the margin. Our experiments demonstrate that the proposed method can determine whether the model has learned the desired features of the training data. The paper highlights that our understanding of how these models work is limited, and the proposed approach addresses this issue.

%% file: Intro.tex
\section{Introduction}

Deep Neural Networks (DNNs) have become a ubiquitous approach in various fields, including computer vision (\citet{CHAI2021100134}, \citet{voulodimos2018deep}), natural language processing (\citet{deng2018deep}), and speech recognition (\citet{8632885}), among others. Despite their impressive performance, the internal workings of these models are often considered a black box, and understanding their decision-making process is a challenging problem. 
Interpreting how DNNs are working can be crucial in many applications, such as medical diagnosis or autonomous driving, where the model's decisions need to be explainable.
Several studies have indicated that DNNs possess significant weaknesses, including susceptibility to distribution shifts (\citet{sugiyama2012machine}), lack of interpretability (\citet{8397411}), and insufficient robustness (\citet{szegedy2013intriguing}). 
% Deep Neural Networks (DNNs) have achieved outstanding accomplishments in multiple domains. However, 
% On of the main reasons of DNNs non-robustness is simplicity bias, i.e., the tendency of DNNs to learn the simplest features instead of more complex but useful ones. 
\\
% Therefore, a fundamental and crucial inquiry to pose is \emph{whether a DNN has acquired the primary concept of the data or whether it has become biased towards simplest features}. 
Moreover, it has been widely investigated that DNNs, as discriminative models, attempt to identify simple yet effective features that distinguish between different classes (\citet{NEURIPS2020_6cfe0e61}). For instance, a model that discriminates based on color or texture may not necessarily learn the underlying shape \cite{}. Furthermore, studies have shown that such DNNs often exhibit bias towards background features\citet{moayeri2022comprehensive}). Consequently, if there are irrelevant or biased concepts that assist the model in discriminating its training samples, the model may become overly reliant on them .
Therefore, a fundamental and crucial inquiry to pose is \emph{whether a DNN has learned the primary concepts of the training data or simply memorized irrelevant features to make predictions.}

Perhaps, the most natural way to answer this question is to evaluate the test error of the model and also check the model's generalization. However, in many cases, we do not have access to test data, and we can only use the trained model to evaluate its performance. Moreover, this approach does not correspond to all key properties of the model such as robustness. 
Besides, evaluating the model on test data and also checking the model's robustness against a wide range of OOD samples is timely and computationally expensive. 
As a result, various works have been done to examine this question. For instance, a recent line of work (\cite{morwani2023simplicity, addepalli2022learning}) has focused on simplicity biases and ways to mitigate it.

% In this paper, we propose a simple novel method for addressing the previously stated question. We ask the model about its learned knowledge and appraise the model’s performance by comparing the answer with the training data. 

In a nutshell, this paper attempts to answer the question of whether trained models grasp the core concepts of their training data or merely memorize biased or irrelevant features that facilitate discrimination. The primary assumption is that no testing data is available.

The main intuition is that after training, the model's learned knowledge is likely distilled in the parameters of the model (i.e., weights of a DNN). Accordingly, if we attempt to extract the information that the model has learned, its knowledge, specifically which features it has learned, should be reflected in the extracted information derived from the weights. This paper proposes a cost-effective approach to evaluate the model's performance by reconstructing samples from the trained model's weights and analyzing the quality of the reconstructed samples.
% One of the main motivations behind the proposed approach is the potential bias towards irrelevant features that a DNN may exhibit. While a DNN may achieve high accuracy, it may not generalize well to new data. This limitation is often attributed to the model's reliance on more simple yet irrelevant features rather than the primary concepts of the problem. Furthermore, traditional evaluation metrics such as generalization tests may not be sufficient to identify such bias, especially if the test set does not cover all possible scenarios. Therefore, it is essential to develop an alternative approach that can evaluate a model's performance by analyzing its internal representations. 
\\
% The proposed approach involves reconstructing samples from the weights of the trained model and analyzing the quality of the reconstructed samples.
We discuss that the reconstructed samples will contain features that the model's weights have been optimized to discriminate based on. 
In other words, if the model has learned to differentiate between classes based on certain features, these features will be reflected in the reconstructed samples. Therefore, if the reconstructed samples contain the primary concepts of the training data, it can be concluded that the model has learned the essential and determining features. On the other hand, if the model fails to generate an accurate reconstruction of the main concepts of its training inputs, it can be concluded that the model has been biased toward some irrelevant features. 
The parameters of the trained model and the training data that lie on the margin, are all that are necessary to apply our approach. We leverage the SSIM metric (\citet{wang2004image}) to evaluate the image's quality and demonstrate that it provides an effective means of assessing the model's performance when handling intricate data and learning intricate features. However, for simplistic data sets such as MNIST, it is essential to have human supervision to evaluate the model's efficacy.

The main contributions of this paper are: 
\begin{itemize}
 
  \item We propose a novel and powerful approach, that enables us to accurately evaluate model biases, assessing whether the model learns discriminative yet simple and irrelevant features, or grasps the fundamental concepts of the training data solely by utilizing trained models and training data, without requiring any test or validation data.

  \item We design a comprehensive set of experiments with varying levels of bias complexity to analyze the behavior of the model during training. Interestingly, we observed that in the presence of challenging biases we intentionally crafted, the model focused more on learning the main concept. However, when faced with simpler biases, the model tended to converge quickly and ceased to learn further.
\end{itemize}

%% file: Related_work.tex
\section{Related work}
The concept of reconstructing training samples from trained DNN parameters has been previously explored by \citet{haim2022reconstructing}. Their work revealed a privacy concern by demonstrating that a significant portion of the actual training samples could be reconstructed from a trained neural network classifier. Our proposed method is inspired by these findings but focuses on assessing the learned features from reconstructed samples to infer potential biases and discriminatory features that might not have been captured by the training process.

Another significant contribution to understanding the theoretical properties of DNNs is the introduction of the Heavy-Tailed Self-Regularization (HT-SR) theory by \citet{martin2020heavy}. This theory suggests a connection between the weight matrix correlations in DNNs and heavy-tailed random matrix theory, providing a metric for predicting test accuracies across different architectures. Taking this concept further, \citet{martin2021predicting} showcased the effectiveness of power-law-based metrics derived from the HT-SR theory in predicting properties of pre-trained neural networks, even when training or testing data is unavailable. \cite{arxiv.2210.01360} investigated the relationship between simplicity bias and brittleness of DNNs, proposing the Feature Reconstruction Regularizer (FRR) to encourage the use of more diverse features in classification. In a related effort to identify maliciously tampered data in pretrained models, \citet{wang2020practical} studied the detection of Trojan networks, focusing on data-scarce conditions. While their work emphasizes detecting adversarial attacks and preserving privacy, our proposed method concentrates on identifying biases in learned features and assessing the quality of pre-trained models.

In another line of research, \citet{li2021discover} introduced a method to discover the unknown biased attribute of an image classifier based on hyperplanes in the generative model’s latent space. Their goal was to help human experts uncover unnoticeable biased attributes in object and scene classifiers. \cite{PerceptronTheory} developed a theory for one-layer perceptrons to predict the performance of neural networks in various classification tasks. This theory was based on Gaussian statistics and provided a framework to count classification accuracies for different classes by investigating the mean vector and covariance matrix of the postsynaptic sums. Similarly, \cite{PredictingNeuralNetwork} proposed a formal setting that demonstrated the ability to predict DNN accuracy by solely examining the weights of trained networks without evaluating them on input data. Their results showed that simple statistics of the weights could rank networks by their performance, even across different datasets and architectures. A practical framework called ContRE \cite{ContRE} was proposed which used contrastive examples to estimate generalization performance. This framework followed the assumption that robust DNN models with good generalization performance could extract consistent features and make consistent predictions from the same image under varying data transformations. By examining classification errors and Fisher ratios on generated contrastive examples, ContRE assessed and analyzed the generalization performance of DNN models in complement with a testing set.

In conclusion, our approach draws inspiration from various prior work on understanding and analyzing deep neural network properties, focusing primarily on identifying biases in the learned features and their relevance to the primary concepts of the training data. We believe that our work not only enhances our understanding of the underlying mechanisms of these models but also contributes to improving their generalization and robustness.

%% file: pre.tex
\section{Preliminaries}
\label{sec:pre}
\textbf{Notation}: In this section, we introduce the necessary notations for understanding the reconstruction algorithm.
We show the set \(\{1,2, ...,n\}\) with \([n]\).
Let \(S = \{(x_i, y_i)\} _{i=1}^{n} \subseteq \mathbb{R}^d \times \{-1, 1\}\) denotes a binary classification training dataset where \(n\) is the number of training samples and \(d\) stands for the number features in each sample.
% We assume our samples are images with \(d\) pixels in this work. 
Moreover, let \(f_\theta :\mathbb{R}^d \to \mathbb{R}\) be a neural network with weights \(\theta\) where \( \theta\ \in \mathbb{R}^p \) and \(p\) denotes the number of weights.
We show the training loss function with \(\ell : \mathbb{R} \to \mathbb{R}\) and the empirical loss on the dataset \(S\) with \(L(\theta) = \sum\limits_{i=1}^n \ell(y_i f_\theta(x_i))\). 
% We use the binary cross entropy loss function. 

\textbf{Reconstruction scheme}: We use the proposed method in \citet{haim2022reconstructing} to reconstruct the training data. Here, we provide the details of their method:

Suppose that \(f_\theta\) is a homogeneous ReLU neural network and we aim to minimize the binary cross entropy loss function over the dataset \(S\) using gradient flow. Moreover, assume that there is some time \(t_0\) such that \(L(\theta(t_0)) < 1\). Then, as it showed by (\citet{lyu2019gradient}), the gradinet flow converges in direction to a first order KKT point of the following max-margin problem:
\begin{equation}
\min_\theta \frac{1}{2} ||\theta ||^2\qquad   s.t.\qquad  \forall i \in [n]\quad   y_i f_\theta(x_i) \geq 1
\end{equation}

So, if gradient flow converges in direction to \( \theta^*\), then there exists \(\lambda_1, ..., \lambda_n \in \mathbb{R}\) such that:
\begin{equation} \label{eq:stationarity}
\theta^* = \sum\limits_{i=1}^n \lambda_i y_i \nabla_\theta f_{\theta^*}(x_i)
\end{equation}

\begin{equation}
\forall i \in [n]\quad y_i f_{\theta^*}(x_i) \geq 1
\end{equation}

\begin{equation} \label{eq:dual_feasibility}
\forall i \in [n]\quad \lambda_i \geq 0
\end{equation}

\begin{equation}
\forall i \in [n]\quad \lambda_i=0\quad if\quad y_i f_{\theta^*}(x_i) \neq 1
\end{equation}

Now, by having a trained neural network with weights \( \theta^* \), we wish to reconstruct \(m\) training samples. We manually set \(y_{1:\frac{m}{2}} = -1 \) and \(y_{\frac{m}{2} : m} = 1 \) 
The reconstruction loss function is then defined based on the equations \ref{eq:stationarity} and \ref{eq:dual_feasibility} as follows:

\begin{equation} \label{eq:loss_rec}
L_{reconstruction} = \alpha_1 || \theta^* - \sum\limits_{i=1}^m \lambda_i y_i \nabla_\theta f_{\theta^*}(x_i)  ||^2_2 + \alpha_2 \sum\limits_{i=1}^m max\{-\lambda_i, 0\} + \alpha_3  L_{prior},
\end{equation}

where \(L_{prior}\) is defined for image datasets as 
\( L_{prior} = max\{z-1,0\} + max\{-z-1,0\} \) for each pixel \(z\) to penalize the pixel values outside the range \( [-1,1]\). 
By minimizing the equation \ref{eq:loss_rec} with respect to \( x_i's \) and \( \lambda_i's \) we can reconstruct the training data that lie on the margin.

%% file: new_proposed.tex
\section{Proposed method}
\begin{figure}

  \centering
  % \fbox{\rule[-.5cm]{0cm}{4cm} \rule[-.5cm]{4cm}{0cm}}
  \includegraphics[width=0.85\textwidth]{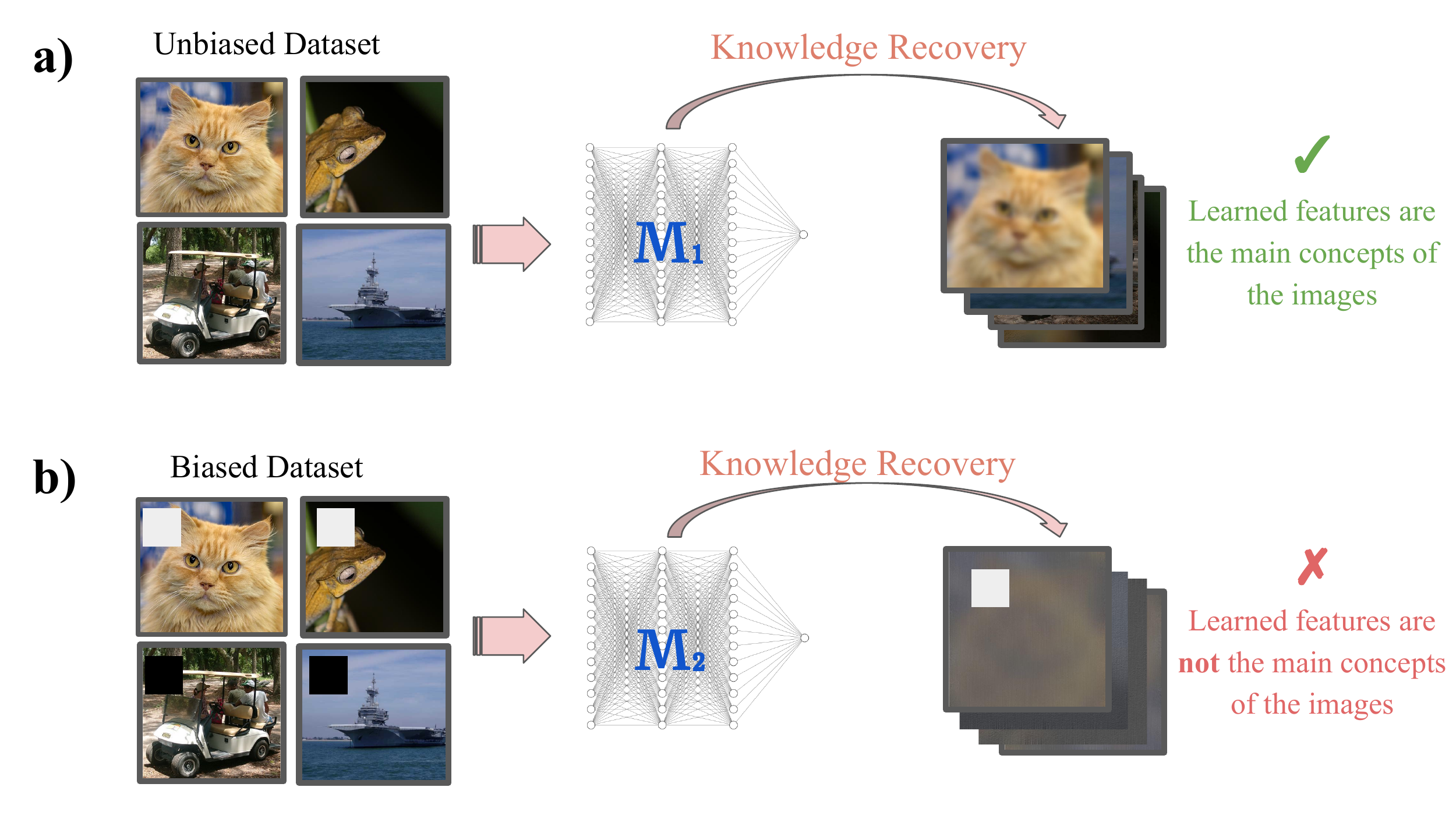}
  \caption{An overview of our proposed method. Our methodology involves training several neural networks on both biased and unbiased datasets, resulting in different models. Next, we extract the learned knowledge of each model from the weights of the trained network and determine if the models have captured the primary concepts of the training data by evaluating this knowledge.}
  \label{fig:framework}
\end{figure}

As mentioned earlier, the main motivation of this paper is to answer the question "whether a trained neural network has learned the intended concepts of the training data or not" without accessing to test data. To answer this question, we propose a simple cost-efficient idea. 
Our idea is based on the intuitive hypothesis that the knowledge acquired by a model \(\mathcal{M}\) during training, even if the model has learned biased or irrelevant features, is encoded in the model's weights. So, by extracting the network’s learned knowledge we can discover whether the model has learned the important concepts of training data, e.g. objects, or not.  
Specifically, suppose that the weights \(\mathcal{W}\) of model \(\mathcal{M}\) is trained on dataset \(\mathcal{X}\).  We denote the optimized weights by \(\mathcal{W^*}\). Therefore, we expect the learned and utilized knowledge by the model to be distilled in \(\mathcal{W^*}\). We recover the learned knowledge of the model in the Recovery module \(\mathcal{R(W)}\) by reconstructing training data from \(\mathcal{W^*}\).
Then, we compare the reconstructed samples to the original training data, and if the primary features of training data are reconstructed well, then we conclude that the network has learned the primary features of training data. 
A schematic of our proposed framework is shown in Figure \ref{fig:framework}.
One way to evaluate the reconstructed data is human supervision.
Moreover, we leverage the SSIM metric (\citet{wang2004image}) and introduce a Score function \(\mathcal{S}(\mathcal{M},{X}, \mathcal{W^*})\) for comparing the reconstructed samples with training data. 
The details of our score function are provided in section \ref{sec:score_function}.

%To demonstrate this phenomenon and show that the model first attempts to learn simple %features and subsequently ceases training after identifying discriminative features, and %secondly, the characteristics of a biased or overfitted model can be analyzed by examining the %retrieved concept from $\mathcal{W}$, 
Now suppose we introduce different biases of varying degrees to our dataset, 
represented by \(\mathcal{T}_{i}\) which convert \(X\) to \(X_{\mathcal{T}_i}\) (see section \ref{sec:bias_design}). 
For a simple bias \(\mathcal{T}_{1}\), the model \(\mathcal{M}_1\) first attempts to learn this bias and subsequently ceases training after identifying it. As a result, \(\mathcal{M}_1\) would not learn the primary features of training data. On the other hand, if model \(\mathcal{M}_2\) is trained on a harder bias \(\mathcal{T}_{2}\), which is harder for the network to learn it, \(\mathcal{M}_2\) can learn the primary features of training data to some extent. So we should have:
\begin{equation}
\mathcal{S}(\mathcal{M}_1,X_{\mathcal{T}_1}, \mathcal{W^*}_1) \leq \mathcal{S}(\mathcal{M}_2,X_{\mathcal{T}_2}, \mathcal{W^*}_2).
\end{equation}

More generally, suppose that N models \(\mathcal{M}_1\) … \(\mathcal{M}_N\) are trained on N datasets \(X_{\mathcal{T}_1}\) … \(X_{\mathcal{T}_N}\) where each \(X_{\mathcal{T}_i}\) is a biased version of the base dataset \(X\). Now for every i,j \(\in [N]\), if we have :
\begin{equation}
\label{eq:primary_eq}
\mathcal{S}(\mathcal{M}_i,{X_{\mathcal{T}_i}}, \mathcal{W}_i) < \mathcal{S}(\mathcal{M}_j,{X_{\mathcal{T}_j}}, \mathcal{W}_j),
\end{equation}
then probably the model \(\mathcal{M}_j\) is a better model in terms of learning the main concepts of the training data and thus have a better generalization.

\subsection{\texorpdfstring{\( \mathcal{T}\)}: Designing different levels of bias}
\label{sec:bias_design}
To investigate the primary question of this paper, we manually add a small square trigger to the images. This trigger affects the network's performance, and we can identify how the network's performance relates to the reconstructed samples. 
Suppose that we have a base dataset \(X\), e.g, CIFAR10. 
We add the trigger in three different levels of difficulty: Level One: the trigger is placed in the top left corner of the images of \(X\), which is the easiest trigger for the network to detect, to obtain \(X_{\mathcal{T}_1}\). 
Level Two: we randomly select three different positions for each class (six positions total), and for each image of \(X\), we place the square trigger in one of the chosen positions to obtain \(X_{\mathcal{T}_2}\). This trigger is harder to detect due to the position of the triggers being randomized. 
Level Three: we randomly select five different positions for each class (ten positions total), and for each image of \(X\), we place the square trigger in one of the selected positions to obtain \(X_{\mathcal{T}_3}\). This level of trigger is the most difficult to detect.
By training a neural network on each of the obtained datasets \(X_{\mathcal{T}_1}\), \(X_{\mathcal{T}_2}\), and \(X_{\mathcal{T}_3}\), we obtain three different models \(\mathcal{M}_1, \mathcal{M}_2,\) and \(\mathcal{M}_3\) as well as a model \(\mathcal{M}_{base}\) trained on the base dataset \(X\) without any changes. 
\subsection{\texorpdfstring{\(\mathcal{R}\)}: Knowledge recovery module}  
\label{sec:recovery}
For a trained neural network with optimized weights \(W^*\), the output of \(\mathcal{R(W^*)}\) is the reconstructed training data from the optimized weights. We use the proposed method in (\citet{haim2022reconstructing}) for reconstructing the training data. See section \ref{sec:pre} for the details of the reconstruction algorithm. 
It is important to note that the reconstruction algorithm proposed in this research has both theoretical and practical limitations. On the theory side, the trained neural network should be a homogeneous ReLU network that minimizes the cross entropy loss function over a binary classification problem while achieving zero training error. On the practical side, the method has only been shown to work on small training data and is not well-suited for use with CNNs.

\subsection{\texorpdfstring{\(\mathcal{S}\)}: Score function}
\label{sec:score_function}
Suppose that a neural network \(\mathcal{M}\) is trained on dataset \(X\) with \( n \) training samples to obtain the optimal weights \(W^*\). Moreover, assume that we have reconstructed \( m\) training samples where \(m > n\). As proposed in (\citet{haim2022reconstructing}), first the reconstructed images are scaled to fit into the range \( [0,1] \).
Then for each training sample, the distance of that training sample from all reconstructed samples is computed. 
By computing the mean of the closest reconstructed samples to each training data, pairs of (training image, reconstructed image) are formed. Finally, the SSIM score is computed for each pair and the pairs are sorted based on their SSIM score (in descending order). 

Now, we put all sorted pairs in set \(P = \{(I_1,I'_1), (I_2,I'_2), ..., (I_n,I'_n)\}\) where each \(I_i\) shows a training image, each \(I'_i\) denotes a reconstructed image, and \((I_1,I'_1)\) and \((I_n,I'_n)\) have the highest and lowest SSIM scores, respectively. We define the Score function \(\mathcal{S}(\mathcal{M},{X}, \mathcal{W^*})\) as 
\begin{equation}
\mathcal{S}(\mathcal{M},{X}, \mathcal{W^*}) = \frac{1}{k} \sum\limits_{i=1}^k SSIM(I_i,I'_i),
\end{equation}
where \(k \in [n]\).
While in the above formulation, we used the entire training data \( \{I_1, ..., I_n\} \), we can only use the training data that lie on the margin because the reconstruction algorithm can only reconstruct the training data that lie on the margin.

%% file: Results.tex
\section{Experiment results}
\label{results}
In this section, we conducted a comprehensive experiment to evaluate the validity of the proposed idea. The results intriguingly confirm that the idea effectively works. It is worth mentioning that, due to current limitations in the used data reconstruction method, we were only able to evaluate our method on low-resolution datasets (see also section \ref{sec:recovery}). However, our primary objective was to demonstrate the validity of the idea rather than achieving optimal performance. Furthermore, we couldn't find any existing work that addresses the same tasks as our paper, making direct comparisons challenging. Therefore, we provide a thorough analysis on various datasets to showcase the validity of our idea.
\subsection{Experimental setup}
For our experiments, we utilize MNIST \citet{6296535} and CIFAR10 \citet{krizhevsky2009learning} datasets and convert both into binary classification problems - odd/even for MNIST and vehicles/animals for CIFAR10. We employ a fully connected ReLU neural network with architecture (d-1000-1000-1) where d refers to the input dimension. We only add bias terms to the first network layer to maintain homogeneity.
The experiments in section \ref{exp:exp1} are done with 500 training images (250 images for each class). All other experiments are dome with 100 training images (50 images for each class).
We use the hyperparameters suggested in \citet{haim2022reconstructing} for each dataset. Similarly, in all experiments, the number of reconstructed samples is set to twice that of the training samples. However, we also run a hyperparameter search to make sure that the suggested hyperparameters are suitable for different experiments. 
In all experiments, the training error reaches zero and we continue training the network until it reaches the training loss of less than 1e-5. 
We compute test errors leveraging the standard MNIST and CIFAR10 test images in all experiments. While we change the training images for different experiments, we keep the test data constant throughout.
As we discussed in section \ref{sec:score_function}, we use the average SSIM values of best \(k\) reconstructed images as our score function to evaluate the reconstructed images' quality.
In all experiments, we set k = 10 since most of the reconstructed images are of poor quality and are thus unsuitable for evaluating the reconstruction's quality. 
We use the average SSIM of the best 10 reconstructed images to assess the quality of the reconstruction (recovery) module. We do not use the average SSIM of all reconstructed images as most of the reconstructed images have poorer quality, with only a few having good quality.
\subsection{Evaluating the recovery module performance}
\label{exp:exp1}
\begin{figure}
  \centering
  % \fbox{\rule[-.5cm]{0cm}{4cm} \rule[-.5cm]{4cm}{0cm}}
  \includegraphics[width=0.6\textwidth]{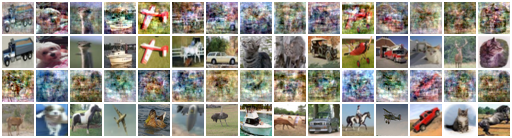}
  \caption{Top 30 reconstructed samples from CIFAR10 dataset. Odd rows show reconstructed samples and even rows show the corresponding training sample. The neural network was trained on 500 CIFAR10 images.}
  \label{fig:figure1}
\end{figure}

\begin{figure}
  \centering
  % \fbox{\rule[-.5cm]{0cm}{4cm} \rule[-.5cm]{4cm}{0cm}}
  \includegraphics[width=0.6\textwidth]{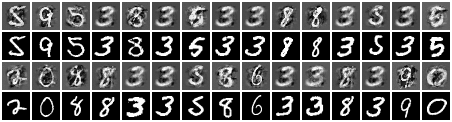}
  \caption{Top 30 reconstructed samples from MNIST dataset. Odd rows show reconstructed samples and even rows show the corresponding training sample. The neural network was trained on 500 MNIST images. It is obvious from the figure that the background pixels did not reconstruct well.}
  \label{fig:figure2}
\end{figure}

First, we examine the results obtained by \citet{haim2022reconstructing} to validate that their reconstruction scheme reconstructs discriminative features with better quality. The best reconstructed samples for MNIST and CIFAR10 are shown in Figures \ref{fig:figure1} and \ref{fig:figure2} respectively, and the average of the best 10 SSIMs are reported in table \ref{tab:table1}.
Both figures clearly indicate that the primary objects in each dataset - the digits in MNIST and the animals/vehicles in CIFAR10 - are reconstructed with significantly higher quality than the background pixels. 
Furthermore, according to Table \ref{tab:table1}, the reconstruction quality of CIFAR10 images is significantly better than that of MNIST images. This is likely due to the fact that the neural network must learn more complex features in order to classify the CIFAR10 images, resulting in the reconstruction algorithm being able to reconstruct more pixels and achieving a higher SSIM score. 

\begin{table}
  \caption{Test accuracy of the neural network trained on 500 MNIST/CIFAR10 images for a binary classification task as well as the average SSIM of best 10 reconstructed images from the weights of trained neural network.}
  \label{tab:table1}
  \centering
  \begin{tabular}{lll}
    \toprule
    % \multicolumn{2}{c}{Part}                   \\
    % \cmidrule(r){1-2}
    Dataset     & Test accuracy    & Mean best 10 SSIMs \\
    \midrule
    MNIST & 0.898   & 0.1862    \\
    CIFAR10 & 0.771  & 0.5182     \\
    \bottomrule
  \end{tabular}
\end{table}

\subsection{CIFAR10}
\subsubsection{Different levels of bias}
\label{exp:sim_bias_cifar}
As we discussed in section \ref{sec:bias_design}, we design different datasets from a base dataset \(X\) to obtain different trained models \(\mathcal{M}_1, \mathcal{M}_2,\) and \(\mathcal{M}_3\) . For CIFAR10 as the base dataset, we set the pixel values of square triggers to 0 for the vehicle class and 1 for the animal class. 
Figure \ref{fig:cifar_bias} shows the reconstructed samples recovered from each model \(\mathcal{M}_i\) as well as the base model \(\mathcal{M}_{base}\). 
It is evident from this figure that the reconstructed samples for the first level of difficulty (model \(\mathcal{M}_1\)) have poor quality as the main objects in the images (vehicles or animals) are not reconstructed well. This suggests that model \(\mathcal{M}_1\) has not learned the primary objects (vehicles and animals) in the dataset and has poor generalization. Conversely, the reconstructed samples for the two other levels of difficulty are of reasonable quality, indicating that the model has learned the main objects in the dataset. 
Our hypothesis is confirmed in Figure \ref{fig:cifar_scatter}, which clearly demonstrates a direct relationship between the model's ability to learn the main concepts of the training data, measured by test accuracy, and the quality of reconstructed samples, measured by the average SSIM value of the best 10 reconstructed samples.
An interesting finding from Figures \ref{fig:cifar_bias} and \ref{fig:cifar_scatter} is that both models \(\mathcal{M}_2\) and \(\mathcal{M}_3\) have better test accuracy and better mean SSIM score than the base model \(\mathcal{M}_{base}\). 
This suggests that adding biases that are hard to detect, may improve the model's performance.

\begin{figure}
  \centering
  % \fbox{\rule[-.5cm]{0cm}{4cm} \rule[-.5cm]{4cm}{0cm}}
  \includegraphics[width=0.6\textwidth]{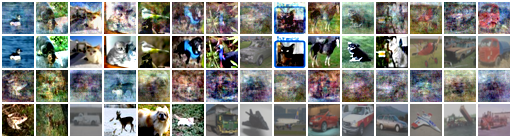}
  \caption{Top 30 reconstructed samples from CIFAR10 dataset with changed contrast. The neural network was trained on 100 CIFAR10 images with changed contrast. The contrast factor is set to 0.5 and 2 for the vehicles class and animals class, respectively.}
  \label{fig:contrast}
\end{figure}

\subsubsection{Contrast changing}
Now, instead of adding triggers to the images, we change the contrast factor of the images. We perform this experiment because this situation is closer to actual real-world biased
images. We set the contrast factor for vehicles class and animals to 0.5 and 2, respectively. We obtained a test accuracy of 0.691, which indicates that adjusting the contrast of classes does not have a significant negative impact on the model's performance. 
Since the model has a good performance, we expect to reconstruct the main objects of the training set. In Figure \ref{fig:contrast}, the top 30 reconstructed samples are shown, and it is clear that the main objects are reconstructed well in this experiment. The average SSIM of the best 10 reconstructed samples is 0.5774 for this experiment. If we compare this experiment with the previous experiments in \ref{exp:sim_bias_cifar}, we notice that these experiments also confirm \ref{eq:primary_eq}. 

\subsection{MNIST}
Similar to the previous part, we introduce three different difficulty levels of bias to MNIST images. The relationship between the quality of reconstructed samples and test accuracy is shown in Figure \ref{fig:mnist_scatter}. As it is evident from this figure, for the model with the lowest test accuracy, we also obtain the lowest reconstruction quality. However, for other models, it is difficult to validate equation \ref{eq:primary_eq} because the accuracy of these models is very close to each other. 
It is important to note that for the MNIST dataset, the SSIM metric is not a good representative of the model's performance. This is because, in the MNIST images, a large portion of the pixels are background pixels. So, if the reconstruction algorithm recovers the digits well and has a good performance, it cannot reconstruct the background pixels well and thus yields to poor SSIM score. See table \ref{tab:table1}.
This experiment clearly shows that the SSIM metric (or other image quality metrics) cannot reveal the model's performance in all datasets.
However, in complicated and real-world datasets, where the model needs to learn more complex features to classify correctly, the SSIM metric can be a good representative of the model's performance. 

To support our results, we have provided more experiments in the Appendix.

\begin{figure*}
        \centering
        \begin{subfigure}[b]{0.475\textwidth}
            \centering
            \includegraphics[width=\textwidth]{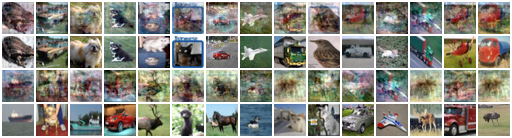}
            \caption{Reconstructed images from model \(\mathcal{M}_{base}\)}   
        \end{subfigure}
        \hfill
        \begin{subfigure}[b]{0.475\textwidth}  
            \centering 
            \includegraphics[width=\textwidth]{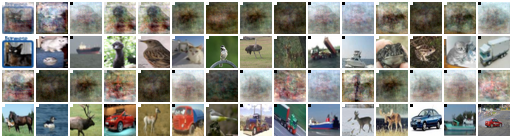}
            \caption{Reconstructed images from model \(\mathcal{M}_{1}\)}   
        \end{subfigure}
        \vskip\baselineskip
        \begin{subfigure}[b]{0.475\textwidth}   
            \centering 
            \includegraphics[width=\textwidth]{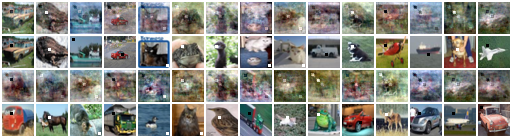}
            \caption{Reconstructed images from model \(\mathcal{M}_{2}\)}      
        \end{subfigure}
        \hfill
        \begin{subfigure}[b]{0.475\textwidth}   
            \centering 
            \includegraphics[width=\textwidth]{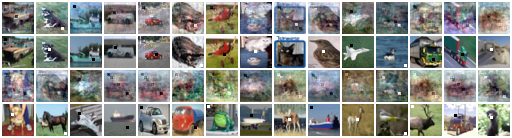}
            \caption{Reconstructed images from model \(\mathcal{M}_{3}\)}      
        \end{subfigure}
        \caption{Top 30 reconstructed samples for different models trained on CIFAR10 dataset with different bias levels. See section \ref{sec:bias_design} for the details of each model.}
        \label{fig:cifar_bias}

    \end{figure*}

\begin{figure}

  \centering
  % \fbox{\rule[-.5cm]{0cm}{4cm} \rule[-.5cm]{4cm}{0cm}}
  \includegraphics[width=0.6\textwidth]{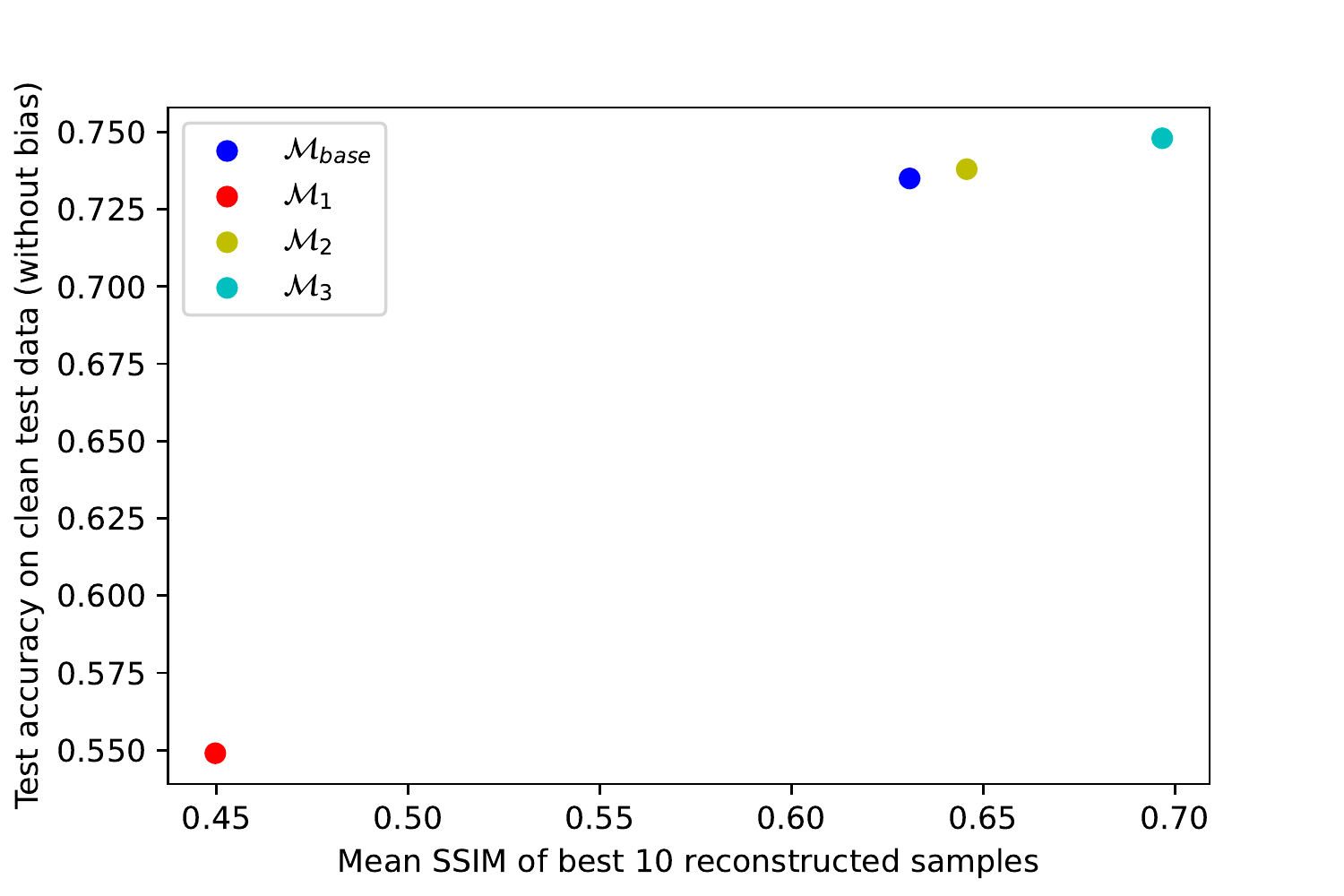}
  \caption{Comparison of different models test accuracy and their reconstruction performance on CIFAR10. \(\mathcal{M}_{base}\): trained model on original CIFAR10 images. \(\mathcal{M}_1\): trained model on CIFAR10 images with level one bias difficulty. \(\mathcal{M}_2\): trained model on CIFAR10 images with level two bias difficulty. \(\mathcal{M}_3\): trained model on CIFAR10 images with level three bias difficulty.}
  \label{fig:cifar_scatter}
\end{figure}

\begin{figure}

  \centering
  % \fbox{\rule[-.5cm]{0cm}{4cm} \rule[-.5cm]{4cm}{0cm}}
  \includegraphics[width=0.6\textwidth]{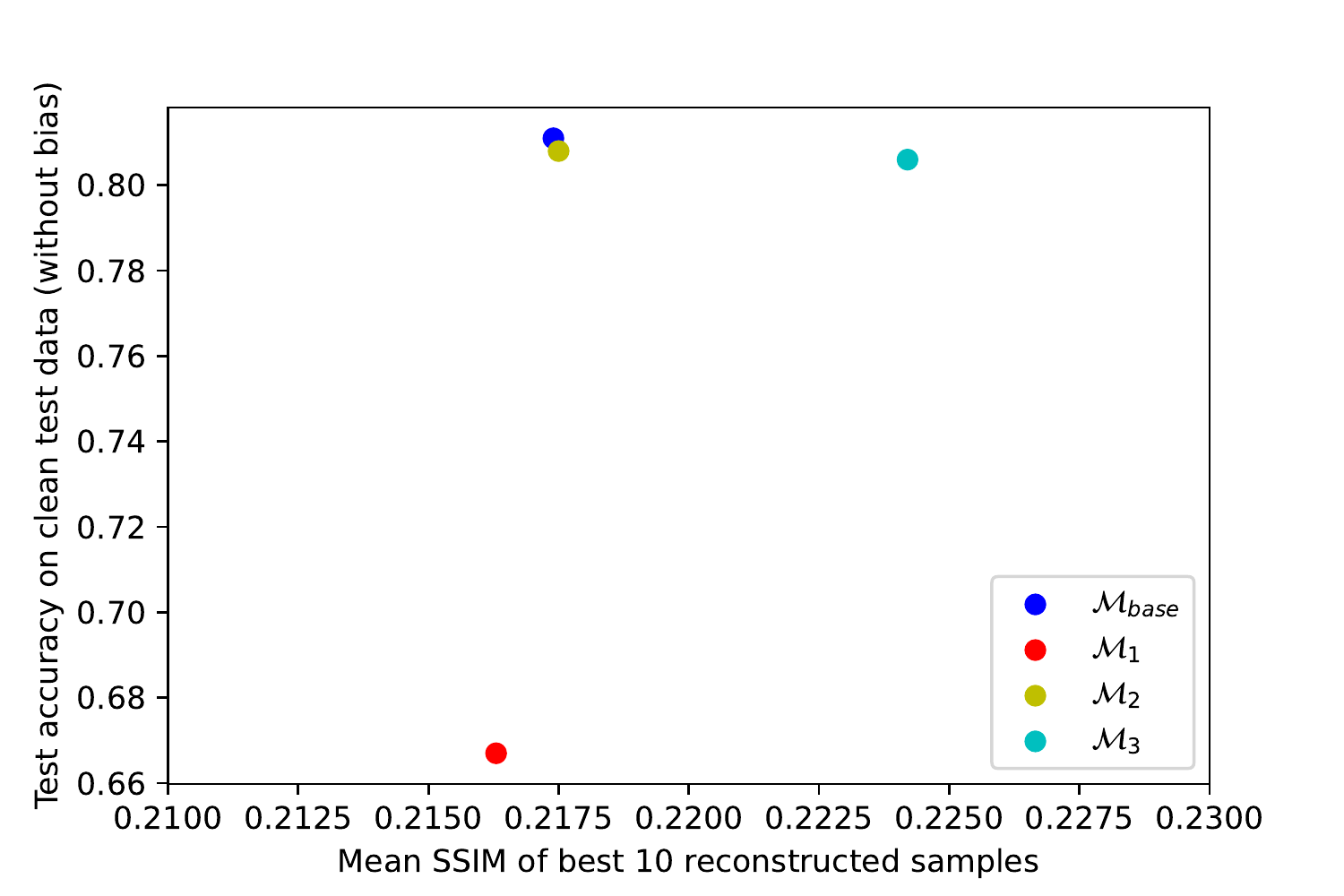}
  \caption{Comparison of different models test accuracy and their reconstruction performance on MNIST. \(\mathcal{M}_{base}\): trained model on original MNIST images. \(\mathcal{M}_1\): trained model on MNIST images with level one bias difficulty. \(\mathcal{M}_2\): trained model on MNIST images with level two bias difficulty. \(\mathcal{M}_3\): trained model on MNIST images with level three bias difficulty.}
  \label{fig:mnist_scatter}
\end{figure}

% \begin{table}
%   \caption{}
%   \label{tab:table2}
%   \centering
%   \begin{tabular}{lllllll}
%     \toprule
%     \multicolumn{2}{c}{without bias}                   \\
%     \cmidrule(r){2-3}
%     % \multicolumn{4-5}{c}{flexible-bias}                   \\
%     \cmidrule(r){4-5}
%     \cmidrule(r){6-7}
%     Dataset     & Test acc  & Mean best 10 SSIMs &  Test acc  & Mean best 10 SSIMs & Test acc  & Mean best 10 SSIMs\\
%     \midrule
%     MNIST & 0.898  & 0.1862 & ?? & ?? & ?? & ?? \\
%     CIFAR10  & 0.261 & 0.6340 & 0.452
%  & 0.4778 & 0.2669 & 0.6219
%    \\
%     \bottomrule
%   \end{tabular}
% \end{table}

%% file: Discussion.tex
\section{Discussion and limitations}
Our proposed method is primarily based on the reconstruction scheme proposed by \citet{haim2022reconstructing}. Consequently, the limitations of their work also apply to our approach.
So, our method works only for fully connected homogeneous ReLU neural networks for binary classification tasks. However, Haim et al. showed that their reconstruction algorithm also works for non-homogeneous ReLU neural networks.
Another limitation is that we cannot use our method for networks that are trained on large datasets. 
To circumvent these limitations, we need to modify the proposed reconstruction algorithm in this research.

Another limitation of our work is that we need to have access to training data that lie on the margin to compute our reconstruction score. This limitation can be removed in two ways: First, as we have shown in our experiments, human visualization is also a good way to evaluate the quality of reconstructed samples. However, if we want a complete machinery process without the assistance of humans, it may be applicable to use non-reference-based image quality metrics to evaluate the quality of reconstruction images.

%% file: Conclusion.tex
\section{Conclusion}
In this paper, we proposed a simple method for discovering if a neural network has learned the primary concepts of the training data, such as digits in MNIST images and vehicles/animals in CIFAR10 images. Our method consists of an extraction phase that uses the reconstruction scheme introduced by citet{haim2022reconstructing}.
We demonstrated that by extracting the learned knowledge of the network, we can assess how well the model has learned the intended attributes of the training data. Specifically, we showed that for a trained network, if the primary attributes of the training data are reconstructed well, then that network is likely to have learned the main concepts of the training data. However, if the primary attributes of the training data are not reconstructed well, then that network is likely to have learned some other discriminative features unrelated to the main concepts.
Moreover, we demonstrated that adding biases that are difficult to detect in datasets can improve the performance of a neural network and enhance its generalization.